\documentclass[runningheads]{llncs}

\usepackage{eccv}

\usepackage{eccvabbrv}

\usepackage{graphicx}
\usepackage{booktabs}
\usepackage[table]{xcolor}
\usepackage{multirow}

\usepackage[accsupp]{axessibility}

\usepackage{hyperref}

\usepackage{orcidlink}

\newcommand{\equalmark}{\textsuperscript{*}}

\newcommand{\corrmark}{\textsuperscript{\textdagger}}

\begin{document}


\title{PhysRAG: Enhancing Physics-Awareness in Video Generation via
Retrieval-Augmented Generation}

\titlerunning{PhysRAG}


\author{
Kexu Cheng\inst{1,2}\equalmark
\and
Zicheng Liu\inst{1}\equalmark
\and
Mingju Gao\inst{1}\equalmark
\\
Chunhe Song\inst{2}\corrmark 
\and 
Hao Tang\inst{1}\corrmark
}


\authorrunning{K. Cheng et al.}



\institute{
School of Computer Science, Peking University, Beijing, China
\and
Institute of AI for Industries, Chinese Academy of Sciences,
Nanjing, China
\\[3pt]
\textsuperscript{*}Equal contribution. \quad
\textsuperscript{\textdagger}Corresponding authors: \email{bjdxtanghao@gmail.com}.
}

\maketitle

\begin{figure*}[t]
    \centering
    \includegraphics[width=0.95\textwidth]{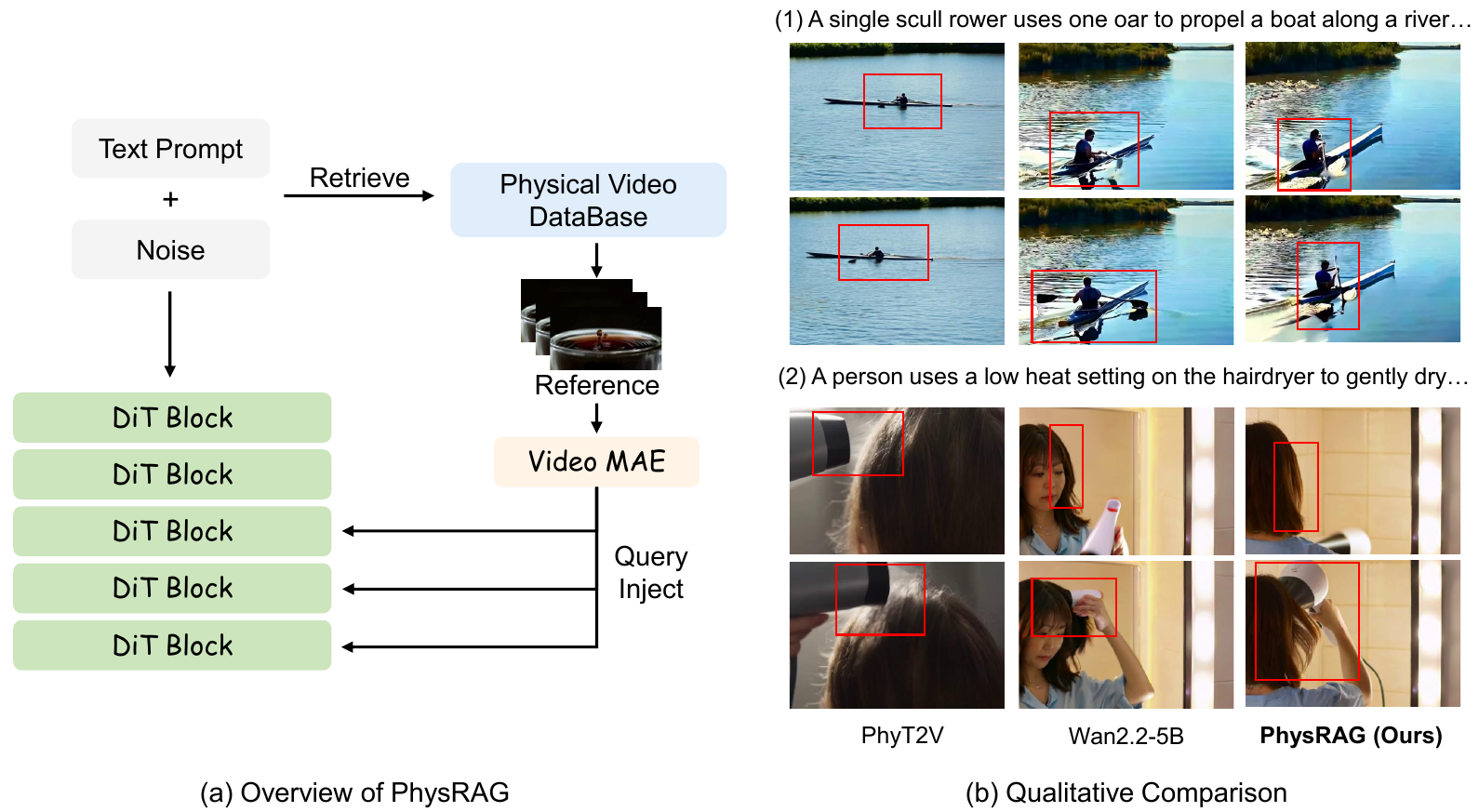}
   \caption{\textbf{Overview and Qualitative Results of PhysRAG.} (a) Architecture of the proposed retrieval-augmented framework. We retrieve relevant physical videos from a database, extract features using Video MAE, and inject these explicit physical priors into the DiT blocks via learnable query injection. (b) Qualitative comparison with baseline models (Wan2.2-5B and PhyT2V). Red boxes highlight regions with complex physical interactions. Compared to the baselines, PhysRAG generates significantly more realistic physical dynamics, such as accurate water-oar interaction (top) and natural hair movement under airflow (bottom).}
    \label{fig:teaser_all}
    \vspace{-0.4cm}
\end{figure*}

\begin{abstract}
Developing physically aware video generation models remains a significant challenge due to the difficulty in capturing diverse physical phenomena, such as thermal dynamics, mechanics, and optics. In this work, we introduce PhysRAG, a novel pipeline that enhances physical awareness in video generation through Retrieval-Augmented Generation (RAG). To address the issue of limited high-quality data, we design a two-stage data filtering pipeline based on the WISA-80K dataset, resulting in a curated set of 7K high-quality videos for training. Furthermore, we construct a physical video database and develop a mechanism to inject physical knowledge into a video diffusion model using learnable queries. Our method achieves state-of-the-art performance in both visual quality and physical rule compliance, surpassing existing models in benchmarks such as PhyGenBench and VBench. We conduct extensive ablation studies to validate the effectiveness of our key components, including the data filtering pipeline, RAG mechanism, and method for physical information extraction. 
To facilitate future research, our code, data, and models are prepared for release at
\url{https://github.com/sediment1024/PhysRAG}.
\end{abstract}



\section{Introduction}
\label{sec:intro}


With the advancement of computing power and the growing scale of training data, text-to-video (T2V) generation has made significant strides~\cite{wan2025wan, yang2024cogvideox, kong2024hunyuanvideo, liu2024sora, wiedemer2025video, gao2025seedance, kling2024}. While the generated visuals appear highly realistic, these models often face difficulties in adhering to physical laws due to the lack of physical modeling. Accurately modeling physical dynamics in generated videos thus remains a challenging and underexplored problem~\cite{kang2024far, liu2025generative, meng2025grounding, bansal2025videophy}.
Overcoming the limitation of physical awareness in video generation models will
enable them to better capture real-world phenomena, which can benefit downstream
applications such as embodied AI~\cite{black2024pi0,intelligence2025pi05visionlanguageactionmodelopenworld, intelligence2025pi06vlalearnsexperience}, robotic manipulation~\cite{wang2025transdiff,liu2024rgbgrasp,huang2025ladi,huang2026paiworld3dconsistentworldfoundation},
autonomous driving~\cite{wang2024drivedreamer,wen2024panacea}, and interactive world modeling~\cite{tong2026scope,ye2026mind}.

A significant body of prior work has focused on addressing physical dynamics in video generation, which can generally be categorized into explicit and implicit approaches. Explicit methods~\cite{zhang2025videorepalearningphysicsvideo, yuan2025newtongen, watters2017visual, toth2019hamiltonian, zhang2025think, wang2025physctrl, xue2025phyt2v, zhao2026phyrpr, zhang2025physchoreo, romero2025learning, battaglia2016interaction, chang2016compositional, guen2020disentangling} rely on deterministic physical simulators or mathematical models to simulate motion directly. While these methods guarantee strict adherence to physical laws, they often struggle with generalization in complex, open-world environments that lie outside their predefined constraints (for example, simulating thermal effects is difficult). On the other hand, implicit methods~\cite{wang2025wisa, ji2025physmaster, liu2025videodpo, cai2025phygdpo, wang2025physcorr, li2025pisa, zhang2026physrvg, hao2025enhancingphysicalplausibilityvideo, chen2025hierarchical, liu2025improving} are predominantly data-driven, focusing on designing physics-aware latent spaces or models to facilitate physically plausible generation. However, these models face a significant challenge in controllability; it remains notoriously difficult to inject precise physical guidance or enforce strict physical constraints within learned implicit representations.

Consider how humans acquire physical knowledge. We perceive and predict events in the physical world by observing scenes in our daily lives, even without consciously understanding the underlying laws of physics. We instinctively anticipate outcomes based on our past experiences with similar physical principles. For instance, we intuitively know that a ball will fall to the ground if we release it, simply by observing freefall. In this process, humans explicitly retrieve relevant scenes from memory (similar to the concept of RAG~\cite{lewis2020retrieval}) and implicitly extract the physical laws governing these scenes, applying them to new scenarios. Similarly, we propose that physical awareness in video generation can be enhanced by explicitly selecting videos that adhere to the same physical laws we want to generate and implicitly injecting the physical principles from these videos into the generation process via learnable queries.

\textbf{Challenge.} Despite the clear motivation, developing a robust video generation model with physical awareness remains challenging. A primary obstacle is the scarcity of high-quality data that captures diverse physical laws, such as thermal dynamics, mechanics, optics, and others. While some datasets aim to train physical-aware video generation models~\cite{wang2025wisa, bear2021physion, tung2023physion++, yi2020clevrercollisioneventsvideo, greff2022kubric, li2025pisa}, these methods either suffer from a lack of high-quality data~\cite{wang2025wisa} or are limited to a small range of physical phenomena, such as freefall~\cite{li2025pisa}. To address this limitation and better train our model, we have designed a two-stage data filtering pipeline. This pipeline first applies a coarse pre-filter to assess the quality of captions and remove those that are irrelevant to the physical phenomenon, retaining the top 10\% of the videos. In the second stage, a fine grounded filter ensures the consistency between video frames and the associated prompts. This approach enables the selection of a high-quality subset that covers a wider range of physical phenomena, providing a more reliable foundation for training our physical-aware video generation model.

Therefore, we propose \textbf{PhysRAG}, a pipeline that enhances physical awareness in video generation through Retrieval-Augmented Generation. To address the issue of scarce high-quality data, we introduce a two-stage data filtering pipeline based on WISA-80K~\cite{wang2025wisa}, which filters out videos with irrelevant captions and low-quality content, resulting in a curated dataset of 7K high-quality videos for training. To make the video diffusion model aware of physics, as illustrated in Figure~\ref{fig:teaser_all} (a), we manually construct a physical video database and extract videos relevant to the given prompts. To inject the physical knowledge from these videos, we design a mechanism that leverages learnable queries to extract and transfer the relevant physical knowledge. As shown in Figure~\ref{fig:teaser_all} (b), our method enables better physical consistency and awareness in video generation.

In our experiment, we evaluate our method on two benchmarks: PhyGenBench~\cite{meng2024towards} and VBench~\cite{huang2024vbench}. Across multiple evaluation settings, our method shows improvements in both visual quality and physical rule compliance, even outperforming closely related models, such as Pika~\cite{pika2024} and Kling~\cite{kling2024}. Additionally, we conduct ablation studies to validate the effectiveness of our proposed modules, including the data filtering pipeline, the RAG mechanism, and the method for extracting and injecting physical information into the video diffusion model. Our contributions can be summarized as follows:

\begin{itemize}
\item We introduce \textbf{PhysRAG}, a novel pipeline that integrates Retrieval-Augmented Generation for enhancing physical awareness in video generation.
\item We design a two-stage data filtering pipeline that significantly improves the quality and relevance of training data, yielding a high-quality curated dataset of 7K videos for training.
\item We construct a manual physical video database and develop a mechanism that leverages learnable queries to inject physical knowledge into the video diffusion model.
\item Our method achieves state-of-the-art performance in physical-aware video generation, demonstrating improvements in both visual quality and rule compliance over existing models.
\end{itemize}


\section{Related Work}
\label{sec:realted_works}

\subsection{Text-to-Video Diffusion Models}
Diffusion-based frameworks have emerged as the dominant paradigm for text-to-video (T2V) generation due to their superior visual quality and scalability. Foundational works established the baseline for temporally coherent synthesis~\cite{ho2022video,singer2022make,ho2022imagen}, while subsequent latent diffusion approaches further enhanced fidelity and training efficiency through efficient adaptation and tuning strategies~\cite{blattmann2023align,chen2024videocrafter2,wang2025lavie,Wu_2023_ICCV}. Recently, the field has converged on large-scale Transformer-based backbones (Video DiT) to improve temporal modeling, as exemplified by systems like Lumiere, Open-Sora, Vchitect-2.0, HunyuanVideo, Wan, CogVideoX, and Step-Video-T2V~\cite{bar2024lumiere,zheng2024open,fan2025vchitect,kong2024hunyuanvideo,wan2025wan,yang2024cogvideox,ma2025stepvideot2vtechnicalreportpractice}. Recent works further explore diffusion and video models as general-purpose visual or world representations, including segmentation, 3D reconstruction, interactive environments, memory consistency, action control, and representation alignment~\cite{wang2026diffusion,wang2025volsplat,tong2026scope,ye2026mind,xiao2026divide}. Despite their rapid progress toward world-simulator capabilities, maintaining physically plausible dynamics and consistent motion over time remains a significant challenge~\cite{brooks2024video,meng2024towards,bansal2025videophy,zhang2025think}.

\subsection{Physics-Aware Video Generation}
Achieving physically plausible dynamics remains a significant challenge in video generation. To address the frequent violation of basic physical laws, existing literature can be broadly categorized into explicit and implicit modeling paradigms. Explicit approaches incorporate physical knowledge through deterministic simulators, trajectory guidance, or structured mathematical constraints~\cite{watters2017visual,toth2019hamiltonian,guen2020disentangling,wang2025wisa,xue2025phyt2v,zhang2025think,yang2025vlipp,wang2025physctrl,zhang2025physchoreo,romero2025learning,hao2025enhancingphysicalplausibilityvideo,zhang2025videorepalearningphysicsvideo}. While ensuring strict adherence to predefined rules, these methods often struggle to generalize across complex, open-world scenarios where phenomena are difficult to explicitly simulate. Conversely, implicit methods are predominantly data-driven, leveraging techniques such as direct preference optimization (DPO), reinforcement learning, 3D constraint reinforcement, and feature alignment to align models with physically plausible distributions~\cite{zhang2025videorepalearningphysicsvideo,liu2025videodpo,liu2025improving,chen2025hierarchical,cai2025phygdpo,wang2025physcorr,li2025pisa,zhang2026physrvg,wang2026world,xiao2026divide}. Although these alignment-based strategies offer broader generalization, they face significant challenges in controllability; because physical priors are typically encoded as scalar rewards or prompt-side reasoning, injecting precise physical guidance into the learned implicit representations remains notoriously difficult.

Despite these advances, existing approaches lack mechanisms to exploit explicit spatiotemporal video references as physical priors. Our method addresses this gap through a retrieval-augmented conditioning design that \textbf{explicitly} retrieves relevant physical exemplars and \textbf{implicitly} injects their encoded dynamics into the latent denoising process via learnable queries.

\subsection{Retrieval-Augmented Conditioning for Video Generation}

Retrieval-augmented generation (RAG) has evolved from text-only external-memory frameworks~\cite{lewis2020retrieval} to multimodal systems like MuRAG and RA-CM3 that jointly retrieve images and text, highlighting the broader utility of non-parametric priors for generation and reasoning~\cite{chen2022murag,yasunaga2022retrieval}. In visual generation, retrieval-conditioned paradigms have further established that external exemplars can improve fidelity and diversity, as demonstrated by retrieval-augmented diffusion and text-to-image generation methods such as RDM and Re-Imagen~\cite{blattmann2022retrieval,chen2022re}.

In the realm of video generation, conditioning strategies are primarily distinguished by the modality of retrieved references and the mechanism of feature injection. Approaches include cross-attention conditioning, parameter-efficient control modules like ControlNet and IP-Adapter~\cite{rombach2022high,zhang2023adding,mou2024t2i,ye2023ip}, and query-based fusion mechanisms such as Q-Former~\cite{NEURIPS2022_960a172b,pmlr-v202-li23q}. Within the video domain, retrieval encoders like VideoCLIP-XL and VideoMAE V2, alongside recent frameworks including RAGME, MotionRAG, and Plug-and-Play Memory, promise exemplar-guided motion transfer~\cite{wang2024videoclip,wang2023videomaev2,10.1145/3731715.3733417,zhu2025motionragmotionretrievalaugmentedimagetovideo,song2025learningplugandplaymemoryguiding}. However, these methods typically target global style or general dynamics. For instance, while Plug-and-Play Memory~\cite{song2025learningplugandplaymemoryguiding} modulates DiTs via frequency-filtered memory tokens, it retrieves from general-domain databases and lacks dedicated mechanisms to extract explicit physical properties, rendering it sub-optimal for physics-grounded generation. Consequently, the query-based \textbf{spatiotemporal} injection of explicit physical priors into the latent denoising process of a Video DiT backbone remains underexplored. Addressing this gap, we introduce a novel retrieval-augmented conditioning design that retrieves physics-relevant video exemplars and injects their encoded spatiotemporal features directly into Video DiT blocks via learnable-query cross-attention and feature-space conditioning.

\section{Physics-Aware Video Data Construction}
\label{sec:data}

\begin{figure*}[t]
    \centering
    \includegraphics[width=\textwidth]{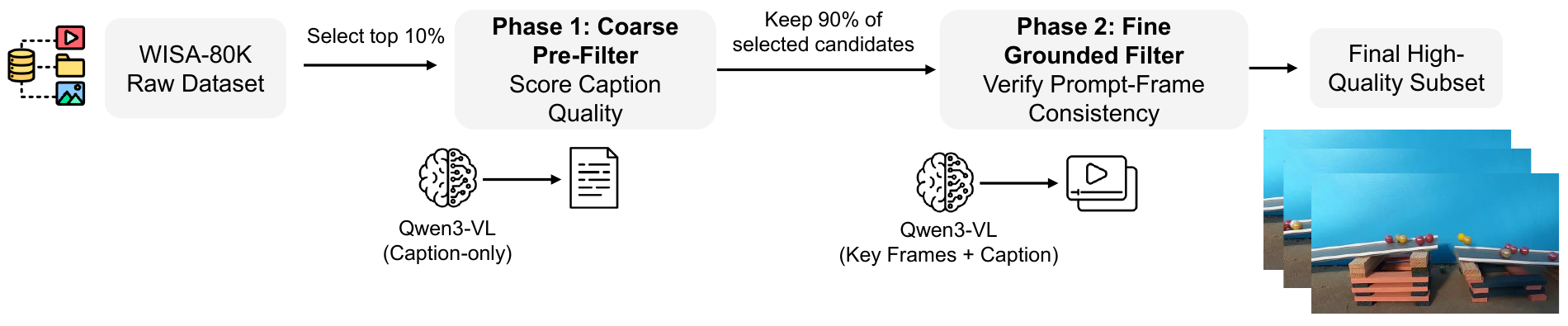}
    \caption{Two-stage preprocessing for WISA-80K. We first use Qwen3-VL (caption-only) to select the 
    top 10\% candidates, then use Qwen3-VL (key frames + caption) to verify prompt--frame consistency
    and retain 90\%, yielding a high-quality subset ($\sim$7K videos).}
    \label{fig:data_preprocessing}
        \vspace{-0.4cm}
\end{figure*}

Constructing a reliable physics-aware video dataset is a foundational prerequisite for training 
trustworthy generative models. Although the existing WISA-80K dataset~\cite{wang2025wisa} provides a 
large collection of internet-sourced videos with coarse-grained cleaning, it cannot be used directly because of persistent data quality issues, as illustrated in Table~\ref{tab:data_phygen}. Such noise can hinder the model’s ability to learn underlying physical laws, making a subsequent cleaning and filtering stage essential.

To address this issue, we introduce a \textbf{two-stage} data preprocessing pipeline to extract a reliable subset from the raw WISA-80K dataset, as illustrated in Fig.~\ref{fig:data_preprocessing}. The pipeline progressively filters out low-quality and physically inconsistent samples, yielding a cleaner and more trustworthy dataset for subsequent model training.

\textbf{Phase 1: Coarse Pre-Filter.} In the first phase, filtering is performed at the text level only. We use Qwen3-VL-4B~\cite{Qwen3-VL} to score the quality and physics relevance of the text descriptions, and retain the top 10\% of candidates. This text-only screening efficiently removes a large fraction of irrelevant samples at low cost, but it inevitably introduces false positives, i.e., samples whose captions appear physically meaningful while the corresponding video content is weakly related or mismatched.

\textbf{Phase 2: Fine Grounded Filter.} To further remove text--video misaligned samples, we introduce a multimodal grounding verification step. For each candidate video retained after Phase 1, we uniformly sample key frames and feed them, together with the corresponding text prompt, into Qwen3-VL-4B~\cite{Qwen3-VL} to assess text--visual consistency. Samples that fail this grounding check are discarded, while the remaining samples are kept for the final subset.

\textbf{Design Rationale.} We adopt a two-phase filtering strategy to balance data quality and computational efficiency. Phase 1 uses low-cost text-only screening to remove a large portion of irrelevant samples, while Phase 2 applies a stricter multimodal grounding check to eliminate false positives from the first stage. In Phase 2, we use uniformly sampled key frames instead of the full video sequence as a cost-effective proxy since processing all frames jointly with the caption using Qwen3-VL-4B is prohibitively expensive at the dataset scale (approximately $2000$ gpu hours on NVIDIA H20 devices). The resulting high-quality subset contains approximately \textbf{7K} videos from the original \textbf{80K} videos in WISA-80K.

\section{The Proposed Method}
\label{sec:method}

\begin{figure*}[t]
    \centering
    \includegraphics[width=\textwidth]{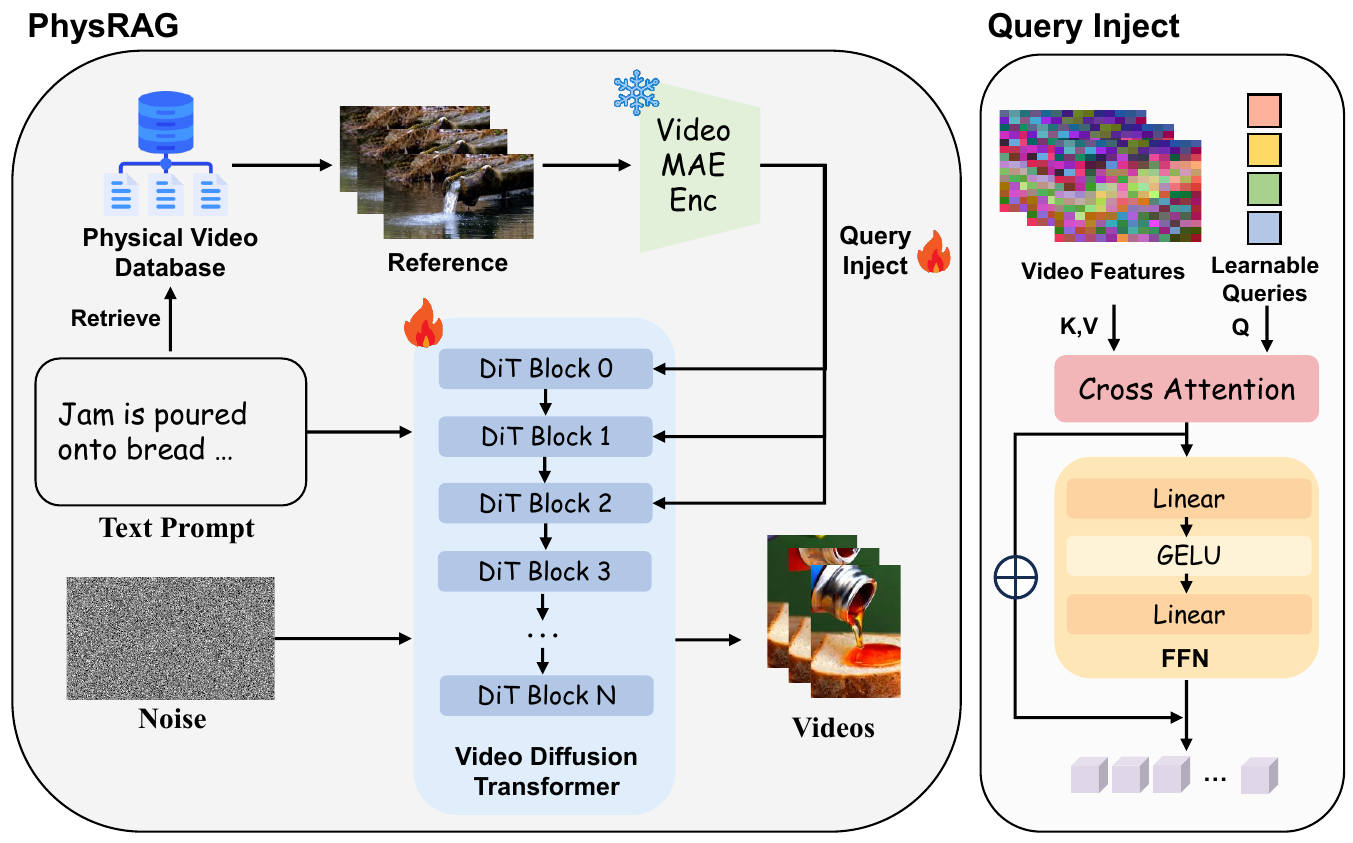}
    \caption{Overview of our retrieval-augmented video diffusion framework. A Video Diffusion Transformer (DiT) generates videos from text and noise while retrieving relevant physical videos from a database. Retrieved features are encoded by Vide MAE and injected into DiT blocks through the Query Inject module.}
    \label{fig:main_pipeline}
        \vspace{-0.4cm}
\end{figure*}

\subsection{Overview}

Figure~\ref{fig:main_pipeline} presents the overall pipeline of our framework. The key idea is to adopt a RAG paradigm that explicitly guides video synthesis with real-world physical dynamics by retrieving reference videos with similar underlying physics from a curated video database.
To this end, we first construct an offline Physical Video Database (Section~\ref{sec:database}), where videos are manually curated and categorized by physical phenomena (e.g., collision, combustion, and explosion). During inference, given a text prompt, we retrieve the most physically relevant video from this database using the VideoCLIP-XL model~\cite{wang2024videoclip}. The retrieved video is then encoded by a pre-trained VideoMAE V2 encoder. The resulting latent representations are injected into the original diffusion transformer blocks via our \textbf{Query Inject} module (Section~\ref{sec:query_inject}), in which learnable query tokens attend to the encoded video latents to capture physically relevant dynamics.

\subsection{PhysRAG Database Construction}
\label{sec:database}

The effectiveness of our retrieval-augmented generation framework critically depends on the quality and organization of the physical video database. As discussed earlier, videos that depict the same physical phenomenon often share similar underlying dynamics. Therefore, constructing a structured database with representative examples of diverse physical phenomena is essential for providing reliable physical priors to guide the generative model.

To this end, we build the \textbf{PhysRAG Database}, a manually curated physical video corpus designed for retrieval at inference time. We first identify common real-world physical phenomena that frequently appear in open-domain video generation scenarios, and systematically group them into \textbf{17 categories}. For each category, we manually collect and curate \textbf{10 high-quality videos} that clearly demonstrate the corresponding physical characteristics, resulting in a database of \textbf{170 videos} in total. Each video is then assigned to its category and stored under the corresponding directory, forming a simple yet effective hierarchical organization for retrieval.
This structured design offers two key benefits. First, it improves retrieval reliability by constraining candidate videos to physically meaningful exemplars. Second, it provides diverse but consistent references within each category, which helps the model extract transferable physical priors instead of overfitting to a single video instance.

\subsection{Physical Prior Injection}
\label{sec:query_inject}



\textbf{Motivation.} In order to explicitly incorporate \textbf{physical priors} into the video generation process, our primary goal is to integrate physical knowledge derived from real-world videos into the Video Diffusion Transformer (DiT). While the pre-trained VideoMAE V2~\cite{wang2023videomaev2} encoder is proficient at extracting rich spatio-temporal features from these retrieved videos, directly embedding such dense latent representations into the DiT presents a significant challenge, as shown in Table~\ref{tab:query_phygen}. This is because our aim is to selectively preserve only the relevant \textbf{physical dynamics} while excluding any irrelevant or extraneous information. If these non-relevant features are incorporated, they could introduce undesirable biases, ultimately degrading the quality of the generated content.

To address this challenge, we introduce the \textbf{Query Inject} module, as depicted in Figure~\ref{fig:main_pipeline}. This module employs a set of learnable queries, which act as an information bottleneck. These queries selectively attend to the video latents, extracting the physical priors that are critical to the task. After the physical priors are distilled, they are injected into the DiT blocks by concatenating them with the original video features. To ensure consistency in the token dimensions of the DiT model, we then use a projector to map the concatenated features back to the original token count. This approach effectively integrates the extracted physical information, enriching the model’s output with essential physical dynamics while avoiding the introduction of irrelevant features or biases. As a result, the generated videos exhibit significantly improved quality and realism.

\textbf{Formulation.}
Given a retrieved physical reference video, we extract its offline VideoMAE-V2
features as $F_v \in \mathbb{R}^{L \times C_v}$, where $L$ is the token sequence
length and $C_v$ is the feature dimension. We further maintain a set of learnable
query tokens $Q \in \mathbb{R}^{N \times C_h}$, where $N=128$ is the number of
query tokens and $C_h$ is the hidden dimension of the query adapter.

The query tokens interact with the physical features through cross-attention:
\begin{equation}
H_{ca} = \mathrm{CrossAttn}(Q, F_v, F_v) \in \mathbb{R}^{N \times C_h},
\end{equation}
where $Q$ serves as the query, while $F_v$ serves as both the key and value.
This operation adaptively aggregates the physical dynamics most relevant to the
current generation process.

We then refine the attended features using a standard feed-forward network with
a GELU activation and a residual connection:
\begin{equation}
H_{ffn} = H_{ca} + \mathrm{FFN}(H_{ca}),
\end{equation}
where
\begin{equation}
\mathrm{FFN}(x) = W_2\,\mathrm{GELU}(W_1 x).
\end{equation}
Afterward, the refined features are projected to a compact physical-prior token
space:
\begin{equation}
H_{out} = H_{ffn} W_o \in \mathbb{R}^{N \times C_p},
\end{equation}
where $C_p$ denotes the physical-prior token dimension, and $W_1$, $W_2$, and
$W_o$ are learnable projection matrices.

Let the intermediate DiT tokens be
$H_{\mathrm{DiT}} \in \mathbb{R}^{T \times C_d}$, where $T$ is the DiT token
length and $C_d$ is the DiT hidden dimension. Before injection, we align the
physical-prior tokens to the DiT token space using a lightweight alignment
operator $\mathcal{A}(\cdot)$, which projects the channel dimension and adjusts
the token sequence to length $T$:
\begin{equation}
\tilde{H}_{p} = \mathcal{A}(H_{out}) \in \mathbb{R}^{T \times C_d}.
\end{equation}

We then concatenate the aligned physical-prior tokens with the DiT tokens along
the sequence dimension and apply a lightweight fusion projector:
\begin{equation}
H_{\mathrm{fuse}} =
\phi\!\left(\mathrm{Concat}(H_{\mathrm{DiT}}, \tilde{H}_{p})\right)
\in \mathbb{R}^{T \times C_d}.
\end{equation}
Finally, we inject the fused physical prior through a gated residual connection:
\begin{equation}
H'_{\mathrm{DiT}} = H_{\mathrm{DiT}} + \alpha H_{\mathrm{fuse}},
\end{equation}
where $\alpha$ is a learnable scalar gate. The resulting $H'_{\mathrm{DiT}}$ is
fed into subsequent DiT blocks, enabling the denoising process to exploit
retrieved physical priors while preserving the original DiT interface.

\section{Experiments and Results}
\subsection{Settings}

\textbf{Data.} Our training data is sourced from WISA-80K~\cite{wang2025wisa}. Following the filtering pipeline outlined in Sec.~\ref{sec:data}, we curate a subset of approximately 7K high-quality videos. This subset is used for all SFT and RAG-based training experiments below.

\textbf{Evaluation.} We evaluate our method on two benchmarks, following~\cite{song2025learningplugandplaymemoryguiding}
\begin{itemize}
    \item \textbf{PhyGenBench}~\cite{meng2024towards} for evaluating physical commonsense, assessed according to the official protocol outlined in \cite{meng2024towards}. The evaluation is conducted using GPT-4o~\cite{hurst2024gpt} as the MLLM judge. PhyGenBench is designed to evaluate the physical reasoning capabilities of models by testing their ability to handle a range of physical scenarios, including mechanics, optics, and other physical phenomena. The protocol measures how well a model predicts and adheres to physical commonsense principles in these contexts, providing a comprehensive benchmark for assessing the integration of physical knowledge into generative models.
    \item \textbf{VBench}~\cite{huang2024vbench}, following the per-dimension protocol on a pre-registered subset that balances temporal and perceptual/semantic factors. In this evaluation, we focus on low-level metrics, such as color, style, object classification, and their average, as well as high-level metrics, including subject consistency, spatial relationships, multi-object handling, and their average.
\end{itemize}

\textbf{Baselines.} We select baselines, including both open-source and closed-source models, to comprehensively evaluate our method. The closed-source commercial models include Pika~\cite{pika2024}, Gen-3~\cite{runway2024gen3}, and Kling~\cite{kling2024}. For open-source models, we compare against prominent text-to-video frameworks such as CogVideoX~\cite{yang2024cogvideox}, Open-Sora~\cite{zheng2024open}, Lavie~\cite{wang2025lavie}, Vchitect 2.0~\cite{fan2025vchitect}, ModelScope~\cite{wang2023modelscope}, and VideoCrafter~\cite{chen2023videocrafter1}. Furthermore, we benchmark against DiT-Mem~\cite{song2025learningplugandplaymemoryguiding}, a recent method that modulates DiTs via frequency-filtered memory tokens. However, since it retrieves from general-domain databases, it lacks explicit physical priors compared to our approach. Finally, Wan 2.2 (5B)~\cite{wan2025wan} serves as our direct baseline to highlight the improvements introduced by our PhysRAG integration.

\textbf{Implementation Details.} We implement our method using PyTorch, building upon the foundational text-to-video (T2V) model Wan2.2-5B~\cite{wan2025wan}. The model is fine-tuned on four NVIDIA H20 GPUs for two days with an effective batch size of 128 (a micro-batch size of 16 per GPU and 2 gradient accumulation steps). We apply a learning rate of $1\times10^{-6}$ and a weight decay of 0.01, training for 20 epochs at a video resolution of $49\times704\times480$. We jointly train the Wan backbone and learnable queries and inject the learnable queries into the 0,1,2 layers of the DiT blocks.
For the PhysRAG module, reference videos are processed using VideoCLIP-XL~\cite{wang2024videoclip} to extract features, while FAISS~\cite{douze2025faisslibrary} is employed for the efficient retrieval of video latents that are similar to the input user prompt. Prior to fine-tuning, we pre-encode videos into VAE latents and prompts into T5 embeddings, which reduces preprocessing overhead and improves training throughput. To optimize memory efficiency, we utilize BF16 mixed precision, gradient checkpointing, and DeepSpeed ZeRO-3 with CPU offloading. We employ the AdamW~\cite{loshchilov2019decoupledweightdecayregularization} optimizer with default parameters ($\beta_1=0.9$, $\beta_2=0.999$) and save checkpoints every 400 steps. 

\subsection{Quantitative Comparisons}

\begin{table*}[t]
\centering
\caption{Quantitative comparison with state-of-the-art video generation models on PhyGenBench. Our proposed PhysRAG achieves the highest overall average score of 0.58, demonstrating superior physical correctness and outperforming both leading closed-source commercial systems and open-source baselines.}
\resizebox{1\textwidth}{!}{%
\begin{tabular}{llcccccc}
\toprule
\textbf{Source} & \textbf{Method} & \textbf{Size} & \textbf{Mechanics}($\uparrow$) & \textbf{Optics}($\uparrow$) & \textbf{Thermal}($\uparrow$) & \textbf{Material}($\uparrow$) & \textbf{Average}($\uparrow$) \\
\midrule
\multirow{3}{*}{Closed} 
& Pika \cite{pika2024} & -- & 0.35 & 0.56 & 0.43 & 0.39 & 0.44 \\
& Gen-3 \cite{runway2024gen3} & -- & 0.45 & 0.57 & 0.49 & \underline{0.51} & 0.51 \\
& Kling \cite{kling2024} & -- & 0.45 & 0.58 & \underline{0.50} & 0.40 & 0.49 \\
\midrule
\multirow{9}{*}{Open} 
& CogVideoX \cite{yang2024cogvideox} & 2B & 0.38 & 0.43 & 0.34 & 0.39 & 0.39 \\
& CogVideoX \cite{yang2024cogvideox} & 5B & 0.39 & 0.55 & 0.40 & 0.42 & 0.45 \\
& Open-Sora V1.2 \cite{zheng2024open} & 1.1B & 0.43 & 0.50 & 0.44 & 0.37 & 0.44 \\
& Lavie \cite{wang2025lavie} & 860M & 0.30 & 0.44 & 0.38 & 0.32 & 0.36 \\
& Vchitect 2.0 \cite{fan2025vchitect} & 2B & 0.41 & 0.56 & 0.44 & 0.37 & 0.45 \\
& DiT-Mem \cite{song2025learningplugandplaymemoryguiding}  & 5B & 0.56 & \textbf{0.74} & 0.48 & 0.47 & \underline{0.56} \\
\cmidrule{2-8}
& Wan 2.2 \cite{wan2025wan} & 5B & \underline{0.58} & 0.60 & \underline{0.50} & 0.48 & 0.54 \\
& \textbf{Wan 2.2 + PhysRAG} & 5B & \textbf{0.59} & \underline{0.66} & \textbf{0.54} & \textbf{0.53} & \textbf{0.58} \\
\bottomrule
\label{tab:physgenbench}
\end{tabular}%
}
\vspace{-0.4cm}
\end{table*}

\begin{table*}[t] 
\setlength{\tabcolsep}{4pt}
\caption{Quantitative evaluation on VBench. Integrating PhysRAG into Wan 2.2 consistently improves both low-level and high-level metrics, achieving the best Low-Avg (65.48\%) and High-Avg (82.88\%) among compared methods.}
\centering
\resizebox{1\textwidth}{!}{%
\begin{tabular}{lcccccccc}
\toprule
\multirow{2}{*}{\textbf{Method}} & \multicolumn{4}{c}{\textbf{Low-level Metrics}} & \multicolumn{4}{c}{\textbf{High-level Metrics}} \\
\cmidrule(lr){2-5} \cmidrule(lr){6-9}
& Color & Style & Obj. Class & Low-Avg & Subj. Consist. & Spatial Rel. & Multi-Obj. & High-Avg \\
\midrule
OpenSora V1.1~\cite{zheng2024open} & 74.56\% & \textbf{23.50}\% & \underline{86.76}\% & 61.61\% & 92.35\% & 52.47\% & 40.97\% & 61.93\% \\
ModelScope~\cite{wang2023modelscope} & 81.72\% & \underline{23.39}\% & 82.25\% & 62.45\% & 89.87\% & 33.68\% & 38.98\% & 54.18\% \\
VideoCrafter~\cite{chen2023videocrafter1} & 78.84\% & 21.57\% & \textbf{87.34}\% & 62.58\% & 86.24\% & 36.74\% & 25.93\% & 49.64\% \\
CogVideo~\cite{hong2022cogvideo} & 79.57\% & 22.01\% & 73.40\% & 58.33\% & 92.19\% & 18.24\% & 18.11\% & 42.85\% \\
DiT-Mem~\cite{song2025learningplugandplaymemoryguiding} & \underline{93.67}\% & 21.22\% & 80.00\% & \underline{64.96}\% & \underline{95.67}\% & \underline{78.42}\% & \underline{74.38}\% & \underline{82.82}\% \\
\midrule
Wan 2.2 \cite{wan2025wan} & 85.92\% & 21.26\% & 79.12\% & 62.10\% & 95.51\% & \textbf{78.91}\% & 69.12\% & 81.18\% \\
\textbf{Wan 2.2 + PhysRAG} & \textbf{93.95}\% & 22.00\% & 80.50\% & \textbf{65.48}\% & \textbf{96.31}\% & 77.33\% & \textbf{75.00}\% & \textbf{82.88}\% \\
\bottomrule
\end{tabular}%
}
\label{tab:vb}
\vspace{-0.4cm}
\end{table*}

\begin{figure*}[t]
    \centering
    \includegraphics[width=0.8\textwidth]{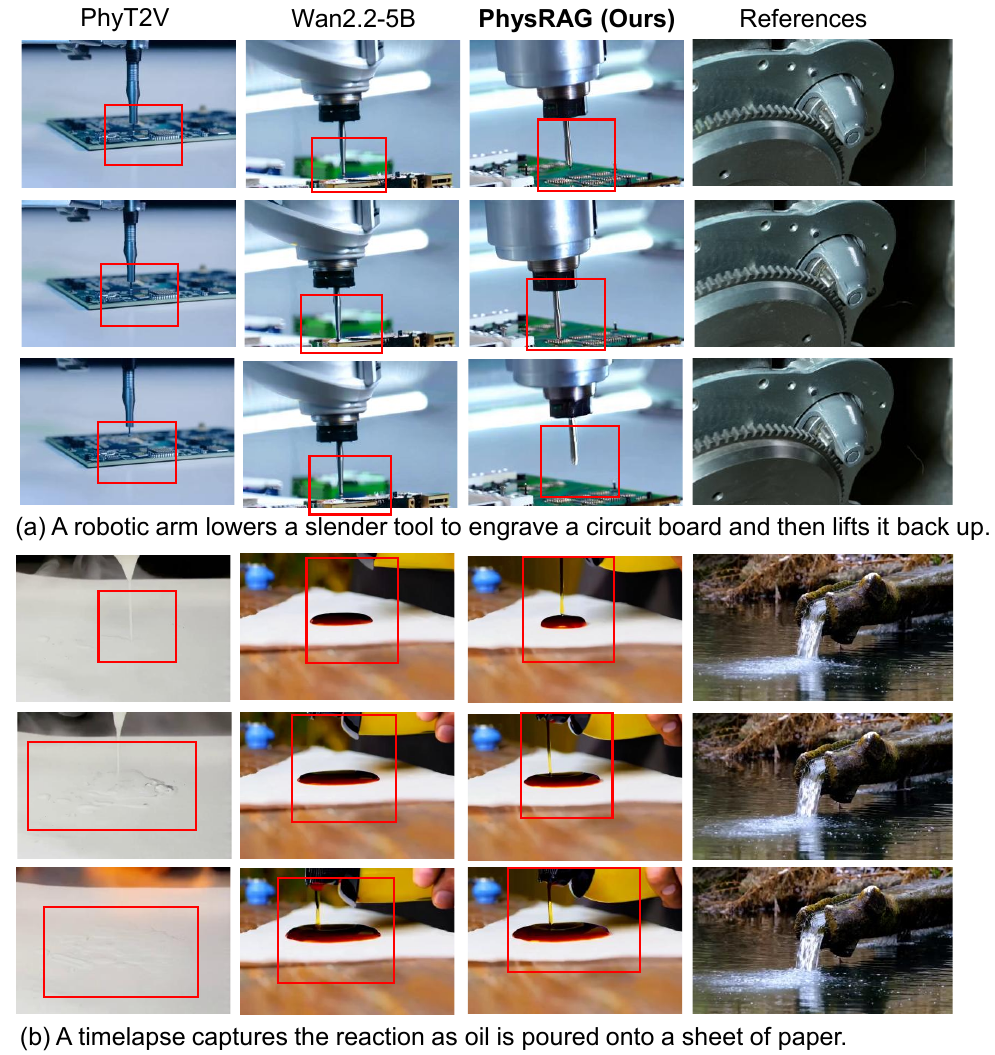}
    \caption{Qualitative comparison of physical consistency across different scenarios from PhyGenBench~\cite{meng2024towards}. PhysRAG outperforms Wan2.2-5B and PhyT2V by accurately modeling physical interactions. In the robotic arm etching scenario, PhysRAG captures the continuous motion required for the task, while baseline models fail to do so. In the fluid dripping example, PhysRAG correctly preserves fluid dynamics and material properties, whereas the baselines exhibit non-causal effects and incorrect behavior.}
    \label{fig:comparison_samples}
    \vspace{-0.4cm}
\end{figure*}

We compare our method with state-of-the-art methods across two comprehensive benchmarks, as shown in Tab.~\ref{tab:physgenbench} (PhyGenBench) and Tab.~\ref{tab:vb} (VBench). Overall, PhysRAG demonstrates superior capability in generating physics-grounded videos while maintaining high visual quality.

\textbf{Results on PhyGenBench.} Tab.~\ref{tab:physgenbench} evaluates the physical correctness of generated videos. Our method achieves the highest average score of \textbf{0.58}, setting a new state-of-the-art that outperforms both leading open-source models and closed-source commercial systems like Kling (0.49). Compared to the foundational Wan-2.2 baseline, PhysRAG brings substantial performance gains in complex physical simulations, with notable absolute improvements of \textbf{0.07} in Thermal and \textbf{0.08} in Material metrics. Furthermore, our approach surpasses the recent memory-guided method DiT-Mem (0.56 average), validating that our explicit physical feature retrieval is more effective for physics-grounded generation than general-domain memory tokens.

\textbf{Results on VBench.} Tab.~\ref{tab:vb} reports the evaluation of general video generation quality and semantic alignment. Integrating PhysRAG into Wan 2.2 consistently improves performance across multiple metrics. In particular, the Low-Avg increases from \textbf{62.10\%} to \textbf{65.48\%}, achieving the best result among all compared models, while the High-Avg improves from 81.18\% to 82.88\%. Our method also achieves the highest Color score of 93.95\% and strong subject consistency and spatial relation performance. These results demonstrate that injecting explicit physical priors into the DiT backbone enhances generation fidelity without compromising semantic alignment.
\begin{figure*}[t]
    \centering
    \includegraphics[width=\textwidth]{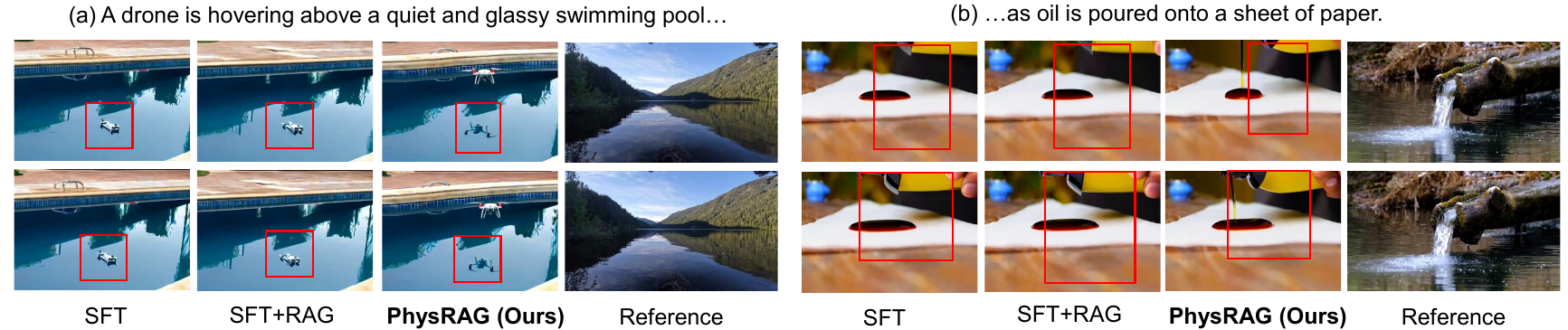}
    \caption{\textbf{Ablation on PhyGenBench~\cite{meng2024towards}.} We compare PhysRAG (ours) with SFT and SFT+RAG on (a) drone-over-pool reflection and (b) oil pouring. In (a), SFT and SFT+RAG miss the water-surface reflection, while PhysRAG preserves it. In (b), SFT and SFT+RAG show non-causal liquid growth without coherent downward flow; PhysRAG produces temporally consistent pouring and accumulation.}
    \label{fig:ablation_study}
\end{figure*}

\subsection{Qualitative Comparisons}

As illustrated in Figure~\ref{fig:comparison_samples}, we qualitatively evaluate the physical consistency of our method against Wan2.2-5B and PhyT2V across different physical scenarios from PhyGenBench~\cite{meng2024towards}. In the first example depicting a robotic arm etching a circuit board, the baseline models fail to comprehend the underlying mechanical interaction, resulting in largely static generations where the tool fails to execute the continuous motion, such as lowering and lifting, required for the etching process. PhysRAG, however, accurately captures the fine-grained, dynamic spatial interaction necessary for this precise mechanical task. In the second example involving fluid dripping onto a surface, the baselines exhibit a severe lack of understanding regarding fluid dynamics and temporal causality. Specifically, Wan2.2-5B generates a non-causal physical effect where the pool of liquid on the surface expands before the falling droplets actually make contact. PhyT2V suffers from a similar unprompted volume increase while also completely failing to render the correct material properties of the fluid. In contrast, PhysRAG successfully preserves the accurate material characteristics and enforces strict causal physics, ensuring that the liquid volume only accumulates sequentially after the fluid physically reaches the surface.

\subsection{Ablation Study}

\begin{table}[t]\small
\centering
\caption{Ablation on Injection Methods.}
\label{tab:query_phygen}
\begin{tabular}{lccccc}
\toprule
Method & Mechanics & Optics & Thermal & Material & Avg \\
\midrule
Concat & \underline{0.541} & \underline{0.633} & 0.488 & 0.500 & 0.540 \\
Cross Attention & \underline{0.541} & \underline{0.633} & \underline{0.522} & \underline{0.516} & \underline{0.553} \\
\textbf{PhysRAG (Ours)} & 
\textbf{0.586} & \textbf{0.660} & \textbf{0.540} & \textbf{0.526} & \textbf{0.578} \\
\bottomrule
\end{tabular}
\vspace{-0.4cm}
\end{table}

\begin{table}[t]
\centering
\caption{Physical prior probing on 17 physical categories.}
\label{tab:prior_probe}
\setlength{\tabcolsep}{5pt}
\renewcommand{\arraystretch}{0.95}
\footnotesize
\begin{tabular}{lcc}
\toprule
Representation & Acc. (\%) & Macro-F1 \\
\midrule
Raw retrieved feat. & 67.65 & 64.25 \\
Adapter tokens & 67.06 & 63.14 \\
\bottomrule
\end{tabular}
\end{table}

\begin{table}[t]\small
\centering
\caption{Ablation on Training Strategies.}
\label{tab:rag_phygen}
\begin{tabular}{lccccc}
\toprule
Method & Mechanics & Optics & Thermal & Material & Avg \\
\hline
SFT & \underline{0.575} & \underline{0.626} & \underline{0.511} & 0.475 & 0.546 \\
SFT+RAG & 0.566 & \underline{0.626} & \underline{0.511} & \underline{0.500} & \underline{0.551} \\
\textbf{PhysRAG (Ours)} & \textbf{0.586} & \textbf{0.660} & \textbf{0.540} & \textbf{0.526} & \textbf{0.578} \\
\bottomrule
\end{tabular}
\vspace{-0.2cm}
\end{table}

\begin{table}[t]\small
\centering
\caption{Ablation of Quality of Training Data.}
\label{tab:data_phygen} 
\begin{tabular}{lccccc}
\toprule
Method & Mechanics & Optics & Thermal & Material & Avg \\
\midrule
Random-SFT & \underline{0.575} & 0.606 & 0.500 & \underline{0.475} & 0.539 \\
Filtered-SFT & \underline{0.575} & 0.626 & \underline{0.511} & \underline{0.475} & \underline{0.546} \\
PhysRAG + Random Data & 0.558 & \underline{0.640} & \underline{0.511} & 0.466 & 0.544 \\
\textbf{PhysRAG + Filtered Data (Ours)} & \textbf{0.586} & \textbf{0.660} & \textbf{0.540} & \textbf{0.526} & \textbf{0.578} \\
\bottomrule
\end{tabular}
\vspace{-0.4cm}
\end{table}

\begin{table}[!htbp] \small
\centering
\caption{Ablation on Query Inject Layers.}
\label{tab:layer_phygen}
\begin{tabular}{lccccc}
\toprule
Method & Mechanics & Optics & Thermal & Material & Avg \\
\hline 
RAG-Front & 0.508 & 0.500 & \underline{0.511} & 0.393 & 0.478 \\
RAG-Mid & 0.533 & \underline{0.653} & 0.500 & \underline{0.500} & \underline{0.546} \\
RAG-Back & \underline{0.541} & 0.580 & 0.500 & 0.464 & 0.521 \\
\textbf{Multi-layer (Ours)} & \textbf{0.586} & \textbf{0.660} & \textbf{0.540} & \textbf{0.526} & \textbf{0.578} \\
\bottomrule
\end{tabular}
\vspace{-0.4cm}
\end{table}

\begin{table}[!htbp]
\centering
\caption{Computational overhead.}
\label{tab:overhead}
\setlength{\tabcolsep}{4pt}
\renewcommand{\arraystretch}{1.0}
\footnotesize
\begin{tabular}{lcccc}
\toprule
Method & Params & Peak Mem. & Latency & Retrieval \\
\midrule
Wan2.2-5B & 4.998B & 19.65GB & 65.68s & -- \\
PhysRAG & 5.114B & 21.46GB & 66.50s & 0.0065s \\
\midrule
Overhead & +2.28\% & +1.81GB & +1.24\% & negligible \\
\bottomrule
\end{tabular}
\end{table}


\textbf{Ablation on Injection Methods.} Table~\ref{tab:query_phygen} presents a comparison of different mechanisms for integrating retrieved features into the Video DiT backbone, evaluating their impact on the generation of physically realistic videos. 
The \textbf{concat} method, which directly concatenates the retrieved features with the video features in the backbone, struggles to handle unaligned spatiotemporal dynamics, yielding the lowest performance with a score of 0.540. 
The \textbf{direct cross attention} method introduces a mechanism that allows the model to attend to the retrieved features while aligning them with the current video features, which improves the performance to a score of 0.553. However, this approach still underperforms due to the introduction of dense reference features that bring significant visual noise, such as irrelevant background details and appearance cues, which hinder the model's ability to extract pure physical priors. In our approach, the \textbf{learnable query-based mechanism} serves as an information bottleneck, selectively focusing on the essential physical dynamics while filtering out disruptive noise. This targeted integration enables the model to explicitly distill the most important physical dynamics, leading to a substantial improvement in performance, with our method achieving a state-of-the-art score of 0.578. 

\textbf{Physical Prior Probing.}
To further verify whether Query Inject preserves physically meaningful
information, we conduct a linear probing experiment on a held-out WISA-80K
probing set with 17 physical categories. As shown in Table~\ref{tab:prior_probe},
the adapter hidden tokens achieve performance close to the raw retrieved
features, with only a small decrease in accuracy and Macro-F1. This suggests
that the learnable-query bottleneck effectively distills physical-category
information from retrieved videos while filtering out irrelevant visual details.

\textbf{Ablation on Training Strategies.} Table~\ref{tab:rag_phygen} and Figure~\ref{fig:ablation_study} evaluate the necessity of joint training. Standard \textbf{SFT} on curated data yields a baseline of 0.546, proving that gains stem significantly from the RAG mechanism rather than just data quality. The \textbf{SFT+RAG} variant (which involves an initial SFT phase, followed by freezing the backbone to train only queries) achieves only 0.551, indicating that joint optimization is crucial for the model to effectively internalize retrieved physical priors. As visualized in Figure~\ref{fig:ablation_study}, \textbf{PhysRAG} with joint training produces superior physical consistency—such as accurate reflections and causal fluid dynamics—compared to frozen-backbone variants. These results validate that the synergy between a trainable DiT backbone and our RAG module is essential for enforcing strict physical laws.


\textbf{Ablation on Quality of Training Data.} Table~\ref{tab:data_phygen} evaluates the effectiveness of our data filtering pipeline by comparing models trained on raw data versus our curated physical dataset. Comparing \textbf{Random-SFT} with \textbf{Filtered-SFT}, the filtered data yields a baseline improvement from 0.539 to 0.546, demonstrating that removing low-quality samples provides a cleaner training signal. More importantly, when integrating our RAG module, \textbf{PhysRAG + Random Data} achieves only 0.544, even underperforming compared to the pure Filtered-SFT baseline. In contrast, our full method, \textbf{PhysRAG + Filtered Data}, reaches a state-of-the-art score of 0.578. 
These results show that RAG benefits substantially from high-quality, physically consistent data.

\textbf{Ablation on Injection Layers.} Table~\ref{tab:layer_phygen} investigates the impact of injecting physical features into different layers of the Video DiT backbone. We compare three single-layer injection variants—\textbf{RAG-Front} (layer 0), \textbf{RAG-Mid} (layer 15), and \textbf{RAG-Back} (layer 24)—against our proposed \textbf{Multi-layer} approach (layers 0, 1, and 2). The results show that single-layer injection, particularly at the extreme front or back, is insufficient to capture the full complexity of physical dynamics. In contrast, our multi-layer strategy achieves the highest performance (0.578) across all metrics.
This superiority stems from two factors: first, injecting into the initial layers allows the DiT to prioritize learning high-level physical representations, early in the generation process; second, the multi-layer design facilitates a more comprehensive integration of physical priors across different levels of feature abstraction, ensuring better consistency in the resulting video.

\textbf{Computational Overhead.}
Table~\ref{tab:overhead} reports the additional cost introduced by PhysRAG
under the same inference setting. Compared with Wan2.2-5B, PhysRAG adds
only 114.25M parameters (+2.28\%) and increases the inference latency by
0.82s per sample (+1.24\%). The FAISS retrieval itself takes only about
0.0065s, which is negligible compared with the diffusion denoising process.
These results indicate that the gains of PhysRAG mainly come from effective
physical-prior retrieval and injection rather than substantially increasing model
scale or inference cost.

\section{Conclusion}
\label{sec:conclusion}

In this work, we presented \textbf{PhysRAG}, a novel retrieval-augmented generation framework designed to enhance physical awareness in video synthesis. By curating a high-quality physical dataset through a two-stage filtering pipeline and developing a learnable query-based injection mechanism, we successfully bridge the gap between explicitly retrieved videos and implicit physics modeling. Our extensive experiments on PhyGenBench and VBench demonstrate that PhysRAG achieves state-of-the-art performance, significantly improving physical consistency in complex scenarios such as fluid dynamics and mechanical interactions while maintaining high visual fidelity. These results highlight the potential of utilizing external physical exemplars to guide diffusion models toward a deeper understanding of real-world physical laws.

\section*{Acknowledgements}

This work was supported by the Fundamental Research Funds for the Central Universities, Peking University.

\clearpage  


%
%
\bibliographystyle{splncs04}
\bibliography{main}
\end{document}